\documentclass[
twocolumn,
]{ceurart}

\usepackage{listings}
\usepackage{multirow}
\usepackage{booktabs}

\usepackage{float}

\begin{document}

\copyrightyear{2025}
\copyrightclause{Copyright for this paper by its authors.
  Use permitted under Creative Commons License Attribution 4.0
  International (CC BY 4.0).}

\conference{CLiC-it 2025: Eleventh Italian Conference on Computational Linguistics, September 24 — 26, 2025, Cagliari, Italy}

\title{MedBench-IT: A Comprehensive Benchmark for Evaluating Large Language Models on Italian Medical Entrance Examinations} 

\author[1]{Ruggero Marino Lazzaroni}
[%
email=ruggero.lazzaroni@uni-graz.at,
url=https://github.com/ruggsea, 
]
\cormark[1] 
\author[2]{Alessandro Angioi}
\author[3]{Michelangelo Puliga}
\author[4]{Davide Sanna}
\author[5]{Roberto Marras}

\cortext[1]{Corresponding author.}
\address[1]{University of Graz} 
\address[5]{OnePix Academy}

\begin{abstract}
Large language models (LLMs) show increasing potential in education, yet benchmarks for non-English languages in specialized domains remain scarce. We introduce MedBench-IT, the first comprehensive benchmark for evaluating LLMs on Italian medical university entrance examinations. Sourced from Edizioni Simone, a leading preparatory materials publisher, MedBench-IT comprises 17,410 expert-written multiple-choice questions across six subjects (Biology, Chemistry, Logic, General Culture, Mathematics, Physics) and three difficulty levels. We evaluated diverse models including proprietary LLMs (GPT-4o, Claude series) and resource-efficient open-source alternatives (<30B parameters) focusing on practical deployability. Beyond accuracy, we conducted rigorous reproducibility tests (88.86\% response consistency, varying by subject), ordering bias analysis (minimal impact), and reasoning prompt evaluation. We also examined correlations between question readability and model performance, finding a statistically significant but small inverse relationship. MedBench-IT provides a crucial resource for Italian NLP community, EdTech developers, and practitioners, offering insights into current capabilities and standardized evaluation methodology for this critical domain.
\end{abstract}


\maketitle

\section{Introduction}
\label{sec:introduction}

Large Language Models (LLMs) have demonstrated remarkable capabilities across diverse tasks \cite{brown2020language}, transforming artificial intelligence applications. Their potential in specialized domains, particularly education \cite{kasneci2023chatgpt,baidoo2023education}, offers promise for personalized learning, assessment, and high-stakes examination support. As LLMs advance, rigorous and contextually relevant evaluation methodologies become essential.

However, a significant portion of existing LLM benchmarks are predominantly English-centric \cite{wang_mmlu-pro_2024,wang_superglue_2020,hendrycks_measuring_2021}, and resources for non-English languages, especially in specific, demanding domains, remain comparatively scarce. This gap is particularly evident for the Italian language, where the lack of specialized benchmarks \cite{attanasio_itaeval_2024,moroni_ita-bench_nodate,attanasio-etal-2024-calamita} can hinder the objective assessment of LLM performance, limit the development of tailored educational technologies, and necessitate reliance on translated materials which may be imperfect or fail to capture local educational nuances.

In this paper, we introduce \textbf{MedBench-IT}, a novel and comprehensive benchmark specifically designed to evaluate the performance of LLMs on Italian medical university entrance examination questions. Sourced from Edizioni Simone, a leading Italian publisher of preparatory materials, MedBench-IT comprises 17,410 expert-written, multiple-choice questions. These questions span six core subjects (Biology, Chemistry, Logic, General Culture, Mathematics, and Physics) and are categorized into three distinct difficulty levels, mirroring the structure of the actual Italian medical admissions tests. Our evaluation encompasses a diverse range of models, including leading proprietary LLMs (e.g., GPT-4o, Claude series) and resource-efficient open-source alternatives (<30B parameters), with a particular focus on models practical for deployment in various Italian organizational contexts.

Our evaluation methodology begins with standard accuracy assessments and is then augmented with several in-depth analyses designed to probe model robustness and behavior. These include rigorous tests for \textbf{reproducibility} (examining response consistency across identical runs), \textbf{ordering bias} (assessing sensitivity to the permutation of answer choices), and the \textbf{impact of explicit reasoning prompts} on model performance. We also investigate the relationship between question text readability and model accuracy, providing further dimensions for understanding model capabilities.

Our primary contributions include:
\begin{itemize}
    \item The creation and presentation of MedBench-IT, the first large-scale benchmark specifically for Italian medical entrance exam questions, curated from expert-validated sources, meant to be a valuable resource for the fostering of LLMs for the Italian language, particularly within its  educational technology sector.
    \item An extensive empirical evaluation of a diverse set of state-of-the-art and practically deployable LLMs on MedBench-IT.
    \item In-depth analyses of model consistency (reproducibility), robustness (ordering bias), and the differential impact of direct versus reasoning-eliciting prompting strategies.
    \item Actionable insights into factors such as subject matter, question difficulty, and text readability that influence LLM performance within this specific Italian educational context.
\end{itemize}

The remainder of this paper is structured as follows: Section~\ref{sec:related_work} discusses related work in LLM evaluation and Italian NLP resources. Section~\ref{sec:benchmark_dataset} details the construction and characteristics of the MedBench-IT dataset. Section~\ref{sec:experimental_setup} outlines our experimental setup, including the core evaluation and subsequent analytical tests. Section~\ref{sec:results_analysis} presents and analyzes the results from these evaluations. Section~\ref{sec:discussion} discusses the broader implications of our findings. Section~\ref{sec:limitations} acknowledges the limitations of our study, and Section~\ref{sec:conclusion} concludes the paper with directions for future work.


\section{Related Work}
\label{sec:related_work}

The evaluation of Large Language Models (LLMs) is a rapidly evolving field, with numerous benchmarks developed to assess their capabilities across various dimensions.

\subsection{General LLM Evaluation Benchmarks}
\label{ssec:general_benchmarks}
Prominent benchmarks such as GLUE (General Language Understanding Evaluation) \cite{wang_glue_2019}, SuperGLUE \cite{wang_superglue_2020}, and MMLU (Massive Multitask Language Understanding) \cite{hendrycks_measuring_2021} have been instrumental in tracking the progress of LLMs on general language understanding and multi-task reasoning. More recently, benchmarks like MMLU-Pro \cite{wang_mmlu-pro_2024} have sought to address saturation issues and increase the challenge level of existing evaluations by incorporating more reasoning-focused questions and more distractor options. While foundational, these benchmarks are predominantly designed for and evaluated in English, limiting their direct applicability to other linguistic contexts without adaptation.

\subsection{Medical Domain LLM Benchmarks}
\label{ssec:medical_benchmarks}
In the medical domain, benchmarks such as MedQA \cite{jin_what_2021}, PubMedQA \cite{jin_pubmedqa_2019}, MedExQA \cite{kim_medexqa_2024}, and challenges like BioASQ \cite{nentidis2023overview} have emerged to evaluate LLMs on medical knowledge, question answering, and reasoning. These resources are crucial for advancing AI in healthcare. However, they primarily focus on English-language materials and examination styles. For example, MedQA \cite{jin_what_2021} is based on USMLE-style questions, which assess medicine-specific knowledge for medical licensing purposes, whereas the Italian medical entrance exam covers a broader range of topics to evaluate candidates' suitability for medical school admission. Applying these benchmarks directly to the Italian medical context presents challenges related to translation fidelity, differences in curriculum emphasis, and distinct examination formats, underscoring the need for native-language, context-specific benchmarks.

\subsection{LLM Evaluation and Resources in Italian}
\label{ssec:italian_benchmarks}
The Italian NLP community has developed evaluation frameworks like the CALAMITA challenge \cite{attanasio-etal-2024-calamita}, which includes the Mult-IT dataset \cite{rinaldi_mult-it_2024} with questions from Italian university entrance and public sector exams. Medical domain efforts include work on specialty tests \cite{casola-etal-2023-testing} and shared tasks like CLinkaRT at EVALITA 2023, focused on the clinical domain \cite{altuna-etal-2023-clinkart}. MedBench-IT distinguishes itself through its specific medical entrance exam focus, a larger specialized corpus (17,410 medical questions), detailed subject/difficulty breakdowns, and comprehensive robustness analyses.

Other evaluation suites for Italian, such as ItaEval \cite{attanasio_itaeval_2024} and ITA-Bench \cite{moroni_ita-bench_nodate}, aim to provide broader assessments of LLM capabilities, often by translating existing English benchmarks or adapting various Italian datasets. In the educational context, benchmarks derived from INVALSI tests (standardized national assessments) like those discussed by Puccetti et al. \cite{puccetti_invalsi_2025} can assess linguistic and mathematical understanding. Unlike these general-purpose benchmarks, MedBench-IT focuses specifically on medical entrance exams using native Italian content.


\subsection{Studies on LLM Robustness and Reasoning}
\label{ssec:robustness_reasoning_studies}
Beyond accuracy, LLM robustness and reasoning capabilities are critical areas of investigation. Prior research has highlighted LLM sensitivity to prompt variations \cite{zhao_calibrate_2021}, ordering biases in multiple-choice questions \cite{wang_mmlu-pro_2024}, and reproducibility challenges. Chain-of-Thought (CoT) prompting \cite{wei_chain--thought_2023} efficacy varies by model and task complexity. Recent work \cite{mirzadeh2024gsmsymbolicunderstandinglimitationsmathematical} revealed significant limitations in LLM mathematical reasoning, showing apparent proficiency may depend more on pattern recognition than genuine understanding. Our work incorporates these considerations by establishing baseline performance and conducting specific experiments to assess reproducibility, ordering bias, and reasoning-eliciting prompt impact on MedBench-IT.


\section{The MedBench-IT Benchmark}
\label{sec:benchmark_dataset}

\subsection{Dataset Construction}
MedBench-IT comprises multiple-choice questions provided by Edizioni Simone, a leading Italian publisher of medical entrance exam preparatory materials. Questions are expert-authored to accurately reflect official Italian medical admission exam style, content, and difficulty.

From an initial corpus of 43,525 questions, we applied filtering steps: (1) removed image-reliant questions for text-based LLM compatibility; (2) excluded English subject questions; (3) stripped XML/HTML markup; (4) standardized format to question stem, five answer options, and single correct answer. After preprocessing, we selected a stratified sample of 17,410 questions maintaining original subject and difficulty proportions, inspired by MMLU's comparable size for balanced coverage and evaluation manageability.

\subsection{Dataset Characteristics and Prompting}
\label{ssec:dataset_characteristics}

The final dataset contains 17,410 questions with metadata indicating subject and difficulty level. Table~\ref{tab:subject_distribution} shows Biology (28.1\%) and Chemistry (22.9\%) as largest portions, followed by Logic (17.3\%), General Culture (13.2\%), Mathematics (9.6\%), and Physics (8.9\%). Table~\ref{tab:difficulty_distribution} shows Level 1/Base (46.1\%), Level 2/Intermediate (41.1\%), and Level 3/Advanced (12.8\%) distributions.

An example Biology question:
\begin{quote}
\small
\textbf{Domanda:} La plasmolisi: \\
\textbf{Possibili risposte:} \\
1. Avviene nelle cellule animali \\
2. E' lo scollamento della membrana plasmatica dalla parete nelle cellule vegetali \\
3. E' causata da un eccessivo turgore della cellula \\
4. Avviene in ambiente ipotonico \\
5. E' la rottura della membrana cellulare nei globuli rossi \\
\textit{(Risposta corretta: 2)}

\textbf{Question (English Translation):} Plasmolysis: \\
\textbf{Possible answers:} \\
1. Occurs in animal cells \\
2. Is the detachment of the plasma membrane from the wall in plant cells \\
3. Is caused by excessive turgor of the cell \\
4. Occurs in a hypotonic environment \\
5. Is the rupture of the cell membrane in red blood cells \\
\textit{(Correct Answer: 2)}
\end{quote}

The distribution of questions by subject is detailed in Table~\ref{tab:subject_distribution}. Biology and Chemistry represent the largest proportions, consistent with the emphasis in typical medical entrance curricula, followed by Logic, General Culture, Mathematics, and Physics.\\
The distribution by difficulty level is presented in Table~\ref{tab:difficulty_distribution}. The majority of questions fall into the base (Level 1) and intermediate (Level 2) categories, with a smaller but significant portion of advanced (Level 3) questions designed to challenge even well-prepared candidates.

\begin{table}[!htbp] 
  \centering
  \caption{Distribution of Questions by Subject in MedBench-IT.}
  \label{tab:subject_distribution}
  \begin{tabular}{lrr}
    \toprule
    Subject           & Count & Percentage (\%) \\
    \midrule
    Biology           & 4,888 & 28.1 \\
    Chemistry         & 3,992 & 22.9 \\
    Logic             & 3,014 & 17.3 \\
    General Culture   & 2,292 & 13.2 \\
    Mathematics       & 1,679 & 9.6  \\
    Physics           & 1,545 & 8.9  \\
    \midrule
    \textbf{Total}    & \textbf{17,410} & \textbf{100.0} \\
    \bottomrule
  \end{tabular}
\end{table}

\begin{table}[!htbp]
  \centering
  \caption{Distribution of Questions by Difficulty Level in MedBench-IT.}
  \label{tab:difficulty_distribution}
  \begin{tabular}{lrr}
    \toprule
    Difficulty Level & Count & Percentage (\%) \\
    \midrule
    Level 1 (Base)       & 8,032 & 46.1 \\
    Level 2 (Intermediate) & 7,153 & 41.1 \\
    Level 3 (Advanced)     & 2,225 & 12.8 \\
    \midrule
    \textbf{Total}       & \textbf{17,410} & \textbf{100.0} \\
    \bottomrule
  \end{tabular}
\end{table}

\subsection{Prompting Strategies}
\label{ssec:prompting_strategies}

To evaluate LLM performance on MedBench-IT, we employed two distinct zero-shot prompting strategies:

\begin{enumerate}
    \item \textbf{Standard Prompt (Direct Answering):} This prompt presents the question and answer choices directly, asking the model to select the number corresponding to the correct answer. The format, presented to the models in Italian, is as follows (see Listing~\ref{lst:standard_prompt}).
    \begin{lstlisting}[language={}, basicstyle=\ttfamily\footnotesize, frame=tb, caption=Standard Prompt Format used in MedBench-IT., label=lst:standard_prompt, escapeinside=||]
Domanda: [testo della domanda]

Possibili risposte:
1. [prima opzione]
2. [seconda opzione]
3. [terza opzione]
4. [quarta opzione]
5. [quinta opzione]

Seleziona il numero della risposta corretta (1-5).
Rispondi nel seguente formato:
Risposta: [numero]
    \end{lstlisting}

    \item \textbf{Reasoning-Eliciting Prompt:} This prompt, used in a separate set of experiments, asks the model to first explain its reasoning process before selecting the correct answer. This is a zero-shot CoT-style prompt \cite{wei_chain--thought_2023}, intended to investigate whether explicitly prompting for reasoning impacts model accuracy. The format is shown in Listing~\ref{lst:reasoning_prompt}.
    \begin{lstlisting}[language={}, basicstyle=\ttfamily\footnotesize, frame=tb, caption=Reasoning-Eliciting Prompt Format used in MedBench-IT., label=lst:reasoning_prompt, escapeinside=||]
Domanda: [testo della domanda]

Possibili risposte:
1. [prima opzione]
2. [seconda opzione]
3. [terza opzione]
4. [quarta opzione]
5. [quinta opzione]

Spiega il tuo ragionamento per arrivare alla risposta e 
poi seleziona il numero della risposta corretta (1-5).
Rispondi nel seguente formato:
Ragionamento: [spiegazione]
Risposta: [numero]
    \end{lstlisting}
\end{enumerate}
The models were instructed to output only the reasoning (if prompted) and the final answer number in the specified format. For experiments utilizing the reasoning prompt, the reasoning text was used for qualitative analysis, while only the numerical answer was used for accuracy scoring.

\section{Experimental Setup}
\label{sec:experimental_setup}

This section outlines the methodology employed for evaluating various Large Language Models (LLMs) on the MedBench-IT benchmark. We detail the models selected for evaluation, the primary metrics used, and the specific protocols for our specialized analyses.

\subsection{Models Evaluated}
\label{ssec:models_evaluated}
A diverse range of LLMs was selected for evaluation on MedBench-IT, encompassing leading proprietary models and prominent open-source alternatives. The selection aimed to provide a comprehensive overview of current model capabilities, including models with specific focus on Italian language tasks and those chosen for practical deployment considerations. For open-source models, we prioritized both state-of-the-art models accessible via API (such as the DeepSeek series) and locally deployable models with parameter counts below 30B. This parameter threshold was established to approximate production environment constraints, specifically targeting models that can be efficiently deployed with less than 40GB of VRAM at half precision.

The proprietary models evaluated included:
\begin{itemize}
    \item OpenAI models accessed via API: the reasoning model o1-preview, GPT-4o, GPT-4 Turbo, GPT-4o mini, and GPT-3.5 Turbo.
    \item Anthropic models accessed via API: Claude 3.5 Sonnet and Claude 3.5 Haiku.
\end{itemize}

The open-source models evaluated represent the latest iterations of various families and sizes at time of experimentation, including several fine-tuned for Italian:
\begin{itemize}
    \item Qwen 2.5 series \cite{qwen2025qwen25technicalreport}: Including instruct versions from 0.5B to 14B parameters (e.g., Qwen 2.5 7B Instruct).
    \item Gemma 2 series \cite{team_gemma_2024}: Including instruct-tuned versions (Gemma 2 2B IT, Gemma 2 9B IT) and community fine-tunes focused on Italian.
    \item Llama 3 series and fine-tunes \cite{grattafiori_llama_2024}: Models such as Llama 3.1 8B Instruct and various Italian fine-tunes contributed by the community.
    \item Phi series \cite{abdin_phi-3_2024}: Including Phi-4.
    \item DeepSeek series \cite{deepseek-ai_deepseek-r1_2025}: Including models accessed via API: DeepSeek Chat (equivalent to Deepseek-V3), DeepSeek Reasoner (equivalent to Deepseek-R1), and locally deployed distilled models (e.g., DeepSeek R1 Distill Qwen 7B).
    \item OLMo 2 series \cite{groeneveld-etal-2024-olmo}: OLMo 2 7B Instruct and OLMo 2 13B Instruct.
    \item Other notable models: Including Aya Expanse 8B \cite{ustun_aya_2024}, and models from the Minerva family by SapienzaNLP \cite{orlando_minerva_nodate}.
\end{itemize}

All open-source models were run locally using standard libraries such as the \textit{vLLM} framework \cite{kwon2023efficient}. For proprietary models, official APIs were used during the experimentation period (between December 2024 and January 2025). Unless specified otherwise (e.g., for reproducibility tests), a sampling temperature of 0 was used for all models to promote deterministic outputs for the main evaluation runs.

\subsection{Evaluation Metrics}
\label{ssec:evaluation_metrics}

The primary metric used for evaluating model performance on MedBench-IT is \textbf{accuracy}. Accuracy is calculated as the percentage of questions for which the model provided the correct answer out of the total number of questions evaluated:
\begin{equation}
\text{Accuracy} = \frac{\text{Number of Correct Answers}}{\text{Total Number of Questions}} \times 100\%
\end{equation}
Accuracy was computed overall, as well as broken down by:
\begin{itemize}
    \item Subject area (Biology, Chemistry, Logic, etc.).
    \item Difficulty level (Level 1, Level 2, Level 3).
\end{itemize}
For the reasoning-eliciting prompt experiments (Section~\ref{ssec:reasoning_impact_test_setup}), only the final numerical answer provided by the model was used to determine correctness for the accuracy calculation; the generated reasoning text itself was used for qualitative observations and length analysis (discussed in Section~\ref{ssec:reasoning_impact_results}).

\subsection{Specialized Analyses Setup}
\label{ssec:specialized_analyses_setup}
In addition to standard accuracy evaluation, we conducted several specialized analyses to assess model robustness and behavior:

\begin{enumerate}
    \item \textbf{Reproducibility Test:}\label{item:reproducibility_test}
    To assess response consistency, we evaluated GPT-4o twice on the entire MedBench-IT dataset using identical parameters (standard prompt, temperature 1). We compared question-by-question responses, calculating percentages of identical answers and consistent correctness across runs (Section~\ref{ssec:reproducibility_results}).

    \item \textbf{Ordering Bias Test:}\label{item:ordering_bias_test}
    To investigate whether answer option order influences predictions, we evaluated selected models (GPT-4o and Claude 3.5 Haiku) on both the original dataset and a version with shuffled answer options, comparing accuracy scores to identify performance deviations attributable to ordering (Section~\ref{ssec:ordering_bias_results}).

    \item \textbf{Reasoning Impact Test:}\label{ssec:reasoning_impact_test_setup}
    All models were evaluated using both standard direct-answering and reasoning-eliciting prompts. Accuracy scores and reasoning text length were analyzed for correlations with answer correctness (Section~\ref{ssec:reasoning_impact_results}).

    \item \textbf{Readability Analysis:}\label{item:readability_analysis}
    We calculated Flesch Reading Ease scores (Formula di Flesch-Vacca) for each question using the `textstat` library\footnote{\url{https://pypi.org/project/textstat/}}. Logistic regression analysis determined whether readability correlates with model performance under both prompt conditions (Section~\ref{ssec:readability_results}).
\end{enumerate}

\section{Results and Analysis}
\label{sec:results_analysis}

This section presents the evaluation results of selected LLMs on MedBench-IT using standard zero-shot prompts, followed by specialized analyses.

\subsection{Overall Model Performance and Subject Difficulty}
\label{ssec:overall_performance}

Table~\ref{tab:main_results} summarizes performance of representative models on MedBench-IT using both standard direct-answering and reasoning-eliciting prompts. Models are grouped into proprietary and open-source categories, with performance reported as overall accuracy (\%).

\begin{table}[!htbp]
  \centering
  \caption{Overall Accuracy (\%) of selected models on MedBench-IT for Standard and Reasoning Prompts.}
  \label{tab:main_results}
  \footnotesize 
  \begin{tabular}{p{2.2cm}ccc}
    \toprule
    Model & Par.\textsuperscript{a} & Std. & Reas. \\
    \midrule
    \multicolumn{4}{l}{\textit{API-based Models}} \\
    \addlinespace[0.5ex]
    DeepSeek-R1 & 671B & 91.9 & 91.8 \\
    o1-preview & -- & 89.1 & 90.7 \\
    Claude 3.5 Son. & -- & 87.8 & 88.3 \\
    DeepSeek Chat & 671B & 86.1 & 87.3 \\
    GPT-4o & -- & 83.9 & 86.8 \\
    GPT-4 Turbo & -- & 83.2 & 79.5 \\
    Claude 3.5 Haiku & -- & 80.4 & 79.8 \\
    GPT-4o mini & -- & 78.7 & 80.9 \\
    GPT-3.5 Turbo & -- & 49.3 & 51.0 \\
    \addlinespace[0.5ex]
    \multicolumn{4}{l}{\textit{Local Models (<30B)}} \\
    \addlinespace[0.5ex]
    Phi-4 & 14B & 76.8 & 67.9 \\
    Qwen 2.5 14B & 14B & 72.6 & 76.9 \\
    Lexora Med. 7B & 7B & 62.1 & 67.2 \\
    Gemma 2 9B & 9B & 61.7 & 69.4 \\
    Qwen 2.5 7B & 7B & 61.1 & 67.6 \\
    Llama 3.1 8B & 8B & 50.3 & 57.4 \\
    Maestrale v0.4 & 7B & 50.8 & 53.0 \\
    Aya Expanse 8B & 8B & 46.7 & 0.1 \\
    Gemma 2 2B & 2B & 41.1 & 34.3 \\
    Qwen 2.5 0.5B & 0.5B & 23.2 & 19.2 \\
    \bottomrule
  \end{tabular}
  \vspace{0.3em}
  \begin{flushleft}
  \scriptsize
  \textsuperscript{a} Par. = Parameters (B = billion, -- = proprietary)
  \end{flushleft}
\end{table}

Top proprietary models and large open-source models like DeepSeek Reasoner and o1-preview achieve accuracy around or above 90\%, followed by Claude 3.5 Sonnet and GPT-4/4o series in the mid-to-high 80s. Open-source models demonstrate strong capabilities, with Phi-4 and Qwen 2.5 14B Instruct achieving 70\%+ accuracy. Models like Gemma 2 9B Instruct, Lexora Medium 7B, and Italian adaptations of Gemma 2 9B (e.g., `anakin87/gemma-2-9b-neogenesis-ita`\footnote{\url{https://huggingface.co/anakin87/gemma-2-9b-neogenesis-ita}}) perform respectably around 60-62\%. Smaller models like Llama 3.1 8B Instruct and the Italian Maestrale family\footnote{\url{https://huggingface.co/mii-llm/maestrale-chat-v0-4-beta}} (based on Mistral 7B) score around 50\%, while many other open-source models, including several Italian fine-tunes of Llama 3 8B, fall into the 30-50\% range. This ranking shows rapid progress in open-source models while still showing a performance delta compared to the best proprietary systems.

Subject analysis reveals consistent difficulty patterns (full per-subject results in Appendix~\ref{app:subject_results}, Table~\ref{tab:subject_results}). \textit{Logic} and \textit{Mathematics} consistently emerge as most challenging for nearly all models. Top models often score 15-25 percentage points lower in Logic compared to Biology or Chemistry (e.g., GPT-4o: 92.4\% in Biology vs 64.9\% in Logic). This suggests abstract reasoning and multi-step problem-solving remain significant hurdles. Conversely, \textit{Biology}, \textit{Chemistry}, and \textit{General Culture} show higher accuracy, likely reflecting strong factual knowledge capabilities. \textit{Physics} performance is typically intermediate.

\subsection{Reproducibility Insights}
\label{ssec:reproducibility_results}

The reproducibility test on GPT-4o yielded 88.86\% response consistency across two identical runs on 17,410 questions, indicating 11.14\% different answer choices despite identical inputs.

\begin{figure*}[!htbp]
  \centering
  \includegraphics[width=0.8\textwidth]{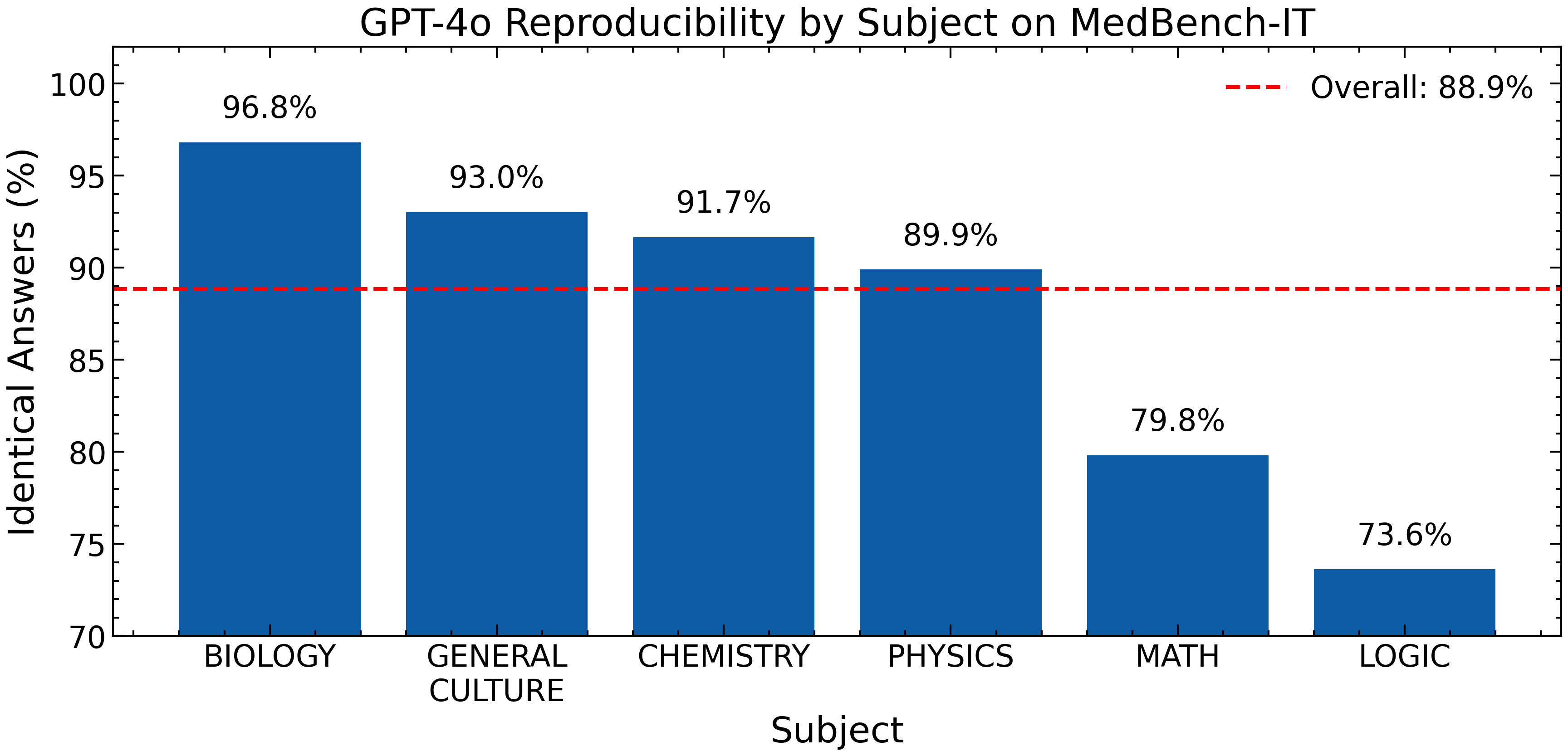}
  \caption{Reproducibility of GPT-4o responses (identical answer choice \%) across different subjects on MedBench-IT.}
  \label{fig:reproducibility_chart}
\end{figure*}

Consistency varied notably across subjects (Figure~\ref{fig:reproducibility_chart}). Higher consistency was observed in knowledge-based subjects like Biology (96.8\%) and General Culture (93.0\%), while lower consistency was found in subjects requiring complex reasoning: Mathematics (79.8\%) and Logic (73.6\%). Physics (89.9\%) and Chemistry (91.7\%) showed intermediate consistency. Across difficulty levels, consistency remained stable (Level 1: 89.8\%, Level 2: 88.1\%, Level 3: 88.0\%).

Regarding correctness, 80.6\% of responses were correct in both runs, 13.2\% were incorrect in both runs, and 6.2\% showed inconsistent correctness between runs. McNemar's test confirmed differences were not statistically significant (p > 0.05), indicating normal stochastic variation rather than systematic instability.

\subsection{Ordering Bias}
\label{ssec:ordering_bias_results}

\begin{figure}[!htbp]
  \centering
  \includegraphics[width=\linewidth]{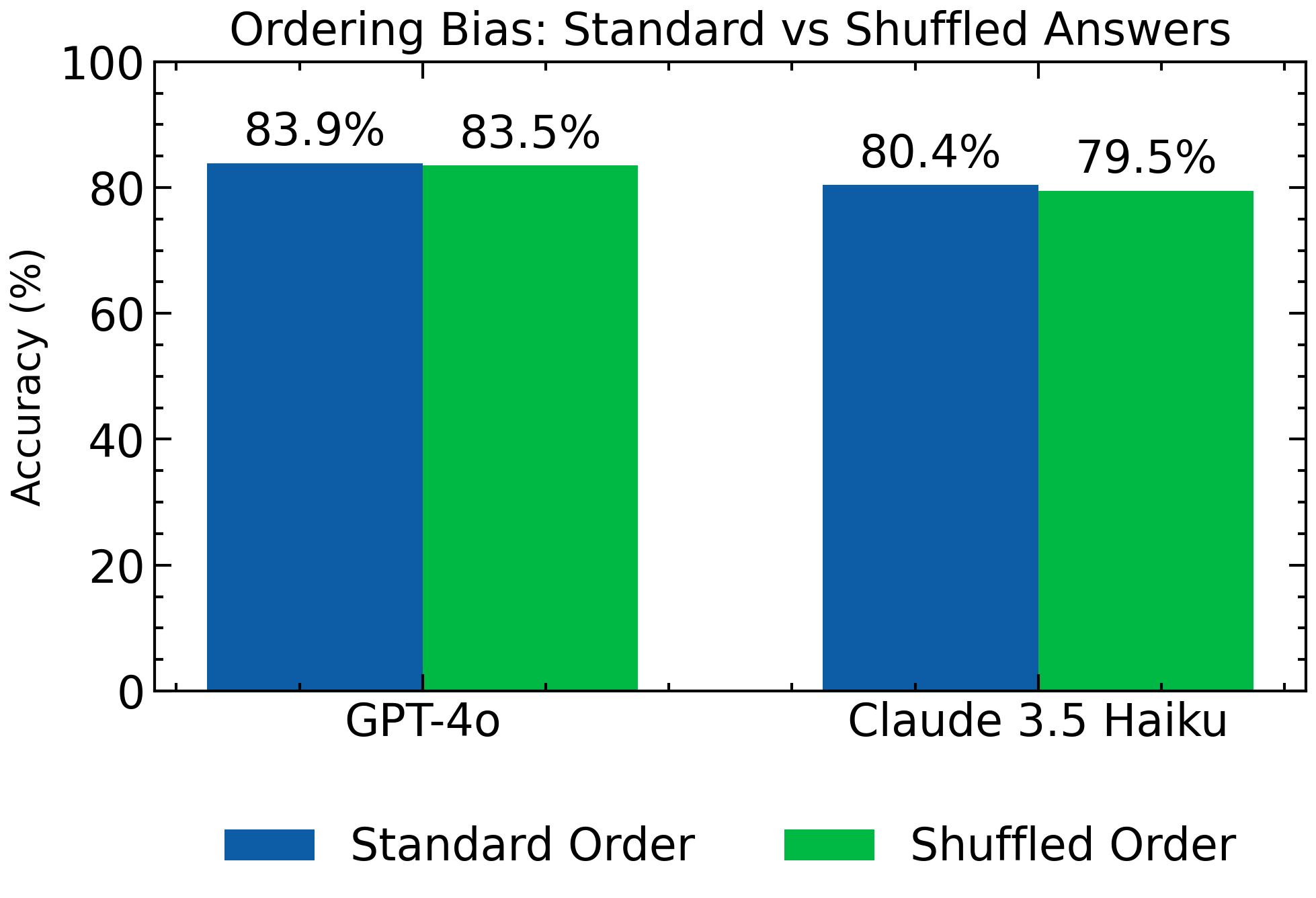}
  \caption{Performance comparison for GPT-4o and Claude 3.5 Haiku on Standard vs. Shuffled MedBench-IT benchmark.}
  \label{fig:ordering_bias_chart}
\end{figure}

The ordering bias test, shuffling answer choices for GPT-4o and Claude 3.5 Haiku, showed minimal impact. GPT-4o's accuracy dropped slightly from 83.9\% to 83.5\% (-0.4\%). Claude 3.5 Haiku decreased from 80.4\% to 79.5\% (-0.9\%) (Figure~\ref{fig:ordering_bias_chart}).

McNemar's test revealed mixed results: GPT-4o showed no statistically significant ordering bias (p > 0.05), while Claude 3.5 Haiku exhibited significant positional sensitivity (p < 0.001). These results demonstrate MedBench-IT's ability to detect ordering bias when present, revealing model-specific robustness differences.

\subsection{Impact of Reasoning Prompts}
\label{ssec:reasoning_impact_results}

\begin{figure*}[!htbp] 
  \centering
  \includegraphics[width=\textwidth]{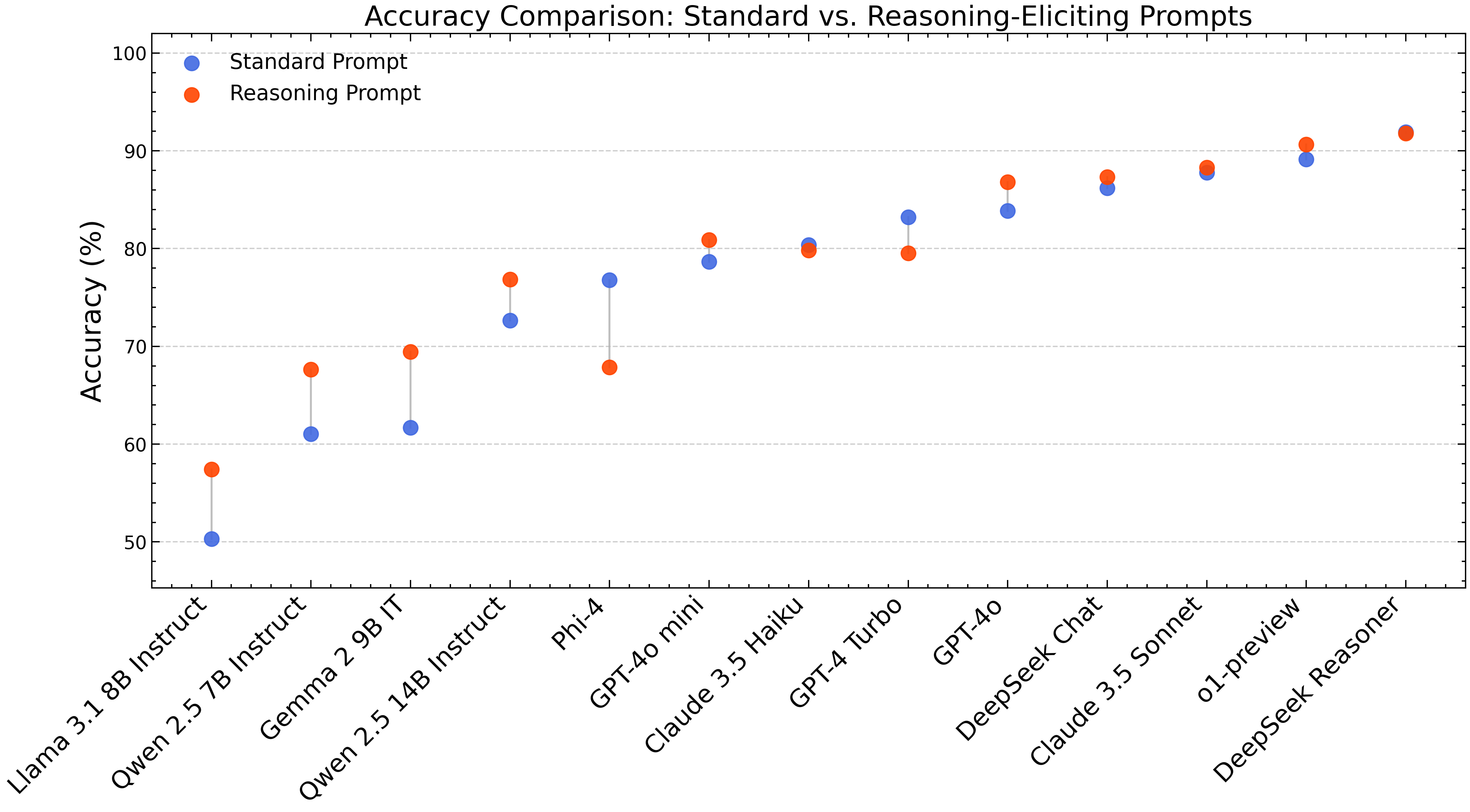}
  \caption{Accuracy comparison for selected models using Standard (blue) vs. Reasoning-Eliciting (red) prompts on MedBench-IT. Models sorted by ascending standard prompt accuracy.}
  \label{fig:reasoning_comparison_chart}
\end{figure*}

Comparing standard direct-answering versus reasoning-eliciting prompts revealed nuanced results (Figure~\ref{fig:reasoning_comparison_chart}). Unlike benchmarks where Chain-of-Thought significantly boosts performance \cite{wei_chain--thought_2023, wang_mmlu-pro_2024}, many top-performing models on MedBench-IT showed no substantial gains, with some exhibiting slightly lower accuracy. Models like DeepSeek Reasoner, o1-preview, and GPT-4o performed slightly worse with reasoning prompts. Some mid-range or smaller models, such as Llama 3.1 8B Instruct, showed slight increases.

This suggests capable models efficiently arrive at answers without requiring explicit, complex reasoning chains. The forced reasoning step might introduce unnecessary processing for some architectures. Analysis showed models tend to produce shorter explanations when correct compared to incorrect answers, indicating more concise justifications for correct answers derived directly.

\subsection{Readability Correlation}
\label{ssec:readability_results}
Analysis investigating the relationship between question text readability (Flesch Reading Ease score for Italian, \textit{Formula di Flesch-Vacca}) and model accuracy revealed a statistically significant, albeit small, inverse correlation. Logistic regression showed lower readability scores (more complex text) were associated with slightly lower odds of correct answers (standard: OR $\approx$ 0.997 per point increase, p < 0.001; reasoning: OR $\approx$ 0.999 per point increase, p < 0.001).

While statistically significant, the small effect size suggests text readability is a minor factor compared to subject knowledge, reasoning complexity, or inherent model capabilities in determining MedBench-IT performance.

\section{Discussion}
\label{sec:discussion}
The evaluation results on MedBench-IT provide several key insights into current LLM capabilities for Italian medical entrance examinations.

The benchmark successfully differentiates performance across models, with top-tier proprietary models (DeepSeek Reasoner, o1-preview, Claude 3.5 Sonnet, GPT-4o) substantially outperforming most open-source alternatives. However, promising mid-sized open-source models (Qwen 2.5 14B, Phi-4) and Italian fine-tunes show competitive results suitable for resource-constrained environments.

Subject-specific analysis reveals Logic and Mathematics as major bottlenecks across all models, suggesting abstract and multi-step reasoning remains challenging compared to knowledge retrieval tasks in Biology or Chemistry. This aligns with observations from other challenging benchmarks.

The reproducibility analysis shows non-negligible variability (11\% response difference, 6\% correctness inconsistency for GPT-4o), particularly in Logic and Mathematics, cautioning against over-interpreting small performance differences on single runs with non-deterministic sampling.

Interestingly, explicit reasoning prompts showed nuanced impact unlike other benchmarks where Chain-of-Thought is essential. Top models often performed slightly worse with reasoning prompts, suggesting they employ efficient internal pathways for these question types. Smaller models showed slight benefits, and shorter reasoning correlated with correctness, indicating potential verbosity when uncertain.

The low correlation with text readability confirms that domain knowledge and reasoning, rather than linguistic complexity, drive difficulty in MedBench-IT.

Overall, MedBench-IT provides a valuable, challenging testbed for the Italian NLP community, highlighting current strengths and weaknesses while supporting evaluation of practical, deployable models for Italian educational applications.

\section{Limitations}
\label{sec:limitations}
While MedBench-IT provides a valuable contribution, several limitations should be acknowledged.

To begin with, the benchmark relies exclusively on a multiple-choice question (MCQ) format, which may not fully capture the depth of understanding compared to open-ended questions. Furthermore, no few-shot evaluation was conducted. This is an interesting extension, particularly for the reasoning approach, where providing complete CoT traces can improve model performance, especially for smaller models. The dataset, while expert-curated, covers preparatory materials and may not fully represent the complexity of advanced medical training. It also does not include context documents, limiting its use for evaluating Retrieval-Augmented Generation (RAG) architectures, which can significantly improve performance.

The potential for data contamination in the pre-training corpora of the evaluated LLMs cannot be entirely ruled out, even if unlikely given our data source. Our robustness analyses were conducted on a limited subset of models, so findings may not generalize. Finally, MedBench-IT is text-only and does not evaluate multimodal reasoning (e.g., interpreting diagrams).

\section{Conclusion and Future Work}
\label{sec:conclusion}

In this paper, we introduced MedBench-IT, the first large-scale benchmark focused on evaluating LLMs on Italian medical university entrance examination questions. By curating 17,410 expert-written questions from a leading publisher, Edizioni Simone, MedBench-IT provides a challenging and contextually relevant testbed spanning six key subjects pertinent to Italian medical admissions.

Our evaluation reveals a clear performance hierarchy. Top proprietary models (DeepSeek Reasoner, o1-preview) achieve near-90\% accuracy, while leading open-source models like Phi-4 and Qwen 2.5 14B exceed 70\%. Italian fine-tunes perform competitively at ~60\%, demonstrating progress in sub-30B parameter models suitable for practical deployment. Logic and Mathematics consistently emerged as the most challenging subjects, indicating complex reasoning remains difficult, while knowledge-intensive subjects like Biology and Chemistry showed higher performance.

Our robustness analyses confirmed ordering bias resistance and good overall reproducibility, though significant variability in Logic and Mathematics emphasizes caution when interpreting complex reasoning results. Explicit reasoning prompts showed nuanced impact—often providing little gain or slight decreases for top models—suggesting MedBench-IT tests applied knowledge and implicit reasoning pathways effectively.

MedBench-IT provides a valuable standardized tool for the Italian NLP community to measure progress, diagnose weaknesses, and evaluate models for Italian EdTech applications.

Future work includes expanding the question set to more advanced medical examinations, conducting deeper qualitative error analysis, and exploring evaluation formats beyond multiple-choice. Furthermore, the complete leaderboard will be hosted and continuously updated on a website maintained by OnePix Academy, allowing for the submission and evaluation of new models.

\begin{acknowledgments} 
 This research was financed by OnePix Academy as part of their effort in Italian EdTech research. We thank Edizioni Simone for providing the dataset used in this benchmark as part of a commercial partnership with OnePix Academy.
\end{acknowledgments}

\section*{Data Availability and Leaderboard} 
\label{sec:data_availability}

Due to the proprietary nature of the source material from Edizioni Simone, the question dataset itself cannot be publicly redistributed. Researchers interested in replicating the benchmark or accessing the data for research purposes should contact the corresponding author to inquire about potential data sharing agreements facilitated through the commercial partnership. As previously mentioned, the complete leaderboard results, including performance metrics for all evaluated models (including those not detailed in the main paper tables/figures) and potentially future model submissions, will be made available and maintained on a dedicated website hosted by OnePix Academy. Interested parties can contact the authors or OnePix Academy for information on submitting new models for evaluation on MedBench-IT.


\bibliography{sample-ceur} 

\appendix 

\section{Additional Question Examples}
\label{app:example_questions}

This appendix provides additional examples of questions from the MedBench-IT dataset, illustrating different subjects. The example for Biology is included in the main text (Subsection~\ref{ssec:dataset_characteristics}). Each question below is presented in Italian, followed by its English translation and the correct answer index.

\subsection{General Culture Example}
\begin{quote}
\small
\textbf{Domanda:} Quale delle seguenti \`e la negazione dell'enunciato "Tutti i bambini amano il gelato''? \\
\textbf{Possibili risposte:} \\
1. [Opzione 1] \\
2. [Opzione 2] \\
3. [Opzione 3] \\
4. [Opzione 4] \\
5. [Opzione 5] \\
\textit{(Risposta corretta: [Index for 'Almeno un bambino non ama il gelato' or similar])}

\vspace{0.5em}
\textbf{Question (English Translation):} Which of the following is the negation of the statement "All children love ice cream"? \\
\textbf{Possible answers:} \\
1. [Option 1] \\
2. [Option 2] \\
3. [Option 3] \\
4. [Option 4] \\
5. [Option 5] \\
\textit{(Correct Answer: [Index for 'At least one child does not love ice cream' or similar])}
\end{quote}

\subsection{Logic Example}
\begin{quote}
\small
\textbf{Domanda:} Se e solo se Giulia a luglio non va in vacanza in montagna, va poi in vacanza al mare ad agosto. Giulia è andata sulle Dolomiti a luglio, dunque non andrà ad agosto al mare. Quale delle seguenti affermazioni segue la stessa struttura logica del suddetto ragionamento? \\
\textbf{Possibili risposte:} \\
1. Carolina, se acquista molte borse, spende molti soldi. Carolina ha acquistato molte borse, dunque ha speso molti soldi \\
2. Clotilde non va in motorino la sera tardi, se piove. Stasera non ha piovuto, dunque è andata in motorino \\
3. Elisa mangia le fragole a cena se e solo se a pranzo non mangia albicocche. Ha già mangiato albicocche a pranzo, dunque a cena non mangia le fragole \\
4. Solo se Clara studia molto, supera gli esami. Clara ha superato gli esami, dunque ha studiato molto \\
5. Se Riccardo non gioca a calcio, non è in forma per giocare a tennis. Riccardo non gioca a tennis, dunque non ha giocato a calcio \\
\textit{(Risposta corretta: 3)}

\vspace{0.5em}
\textbf{Question (English Translation):} If and only if Giulia does not go on holiday to the mountains in July, she then goes on holiday to the sea in August. Giulia went to the Dolomites in July, therefore she will not go to the sea in August. Which of the following statements follows the same logical structure as the reasoning above? \\
\textbf{Possible answers:} \\
1. Carolina, if she buys many bags, spends a lot of money. Carolina bought many bags, therefore she spent a lot of money \\
2. Clotilde does not ride her scooter late at night if it rains. Tonight it did not rain, therefore she went on her scooter \\
3. Elisa eats strawberries for dinner if and only if she does not eat apricots for lunch. She already ate apricots for lunch, therefore she does not eat strawberries for dinner \\
4. Only if Clara studies hard, does she pass the exams. Clara passed the exams, therefore she studied hard \\
5. If Riccardo does not play football, he is not fit to play tennis. Riccardo does not play tennis, therefore he did not play football \\
\textit{(Correct Answer: 3)}
\end{quote}

\subsection{Physics Example}
\begin{quote}
\small
\textbf{Domanda:} In quale sistema una tonnellata è un multiplo? \\
\textbf{Possibili risposte:} \\
1. Nel sistema delle dozzine \\
2. Nel sistema binario \\
3. Nel sistema esadecimale \\
4. Nel sistema decimale \\
5. Nessuna delle altre \\
\textit{(Risposta corretta: 4)}

\vspace{0.5em}
\textbf{Question (English Translation):} In which system is a ton (tonne) a multiple? \\ 
\textbf{Possible answers:} \\
1. In the duodecimal system (base 12) \\
2. In the binary system \\
3. In the hexadecimal system \\
4. In the decimal system \\
5. None of the others \\
\textit{(Correct Answer: 4)}
\end{quote}

\subsection{Chemistry Example}
\begin{quote}
\small
\textbf{Domanda:} A quante moli corrispondono 5 mL (d=1,8 g$\cdot$cm$^{-3}$) di un composto avente una massa molare di 450 g$\cdot$mol$^{-1}$? \\ 
\textbf{Possibili risposte:} \\
1. [Option 1 - e.g., 0.01 mol] \\
2. [Option 2 - e.g., 0.02 mol] \\
3. [Option 3 - e.g., 0.04 mol] \\
4. [Option 4 - e.g., 0.1 mol] \\
5. [Option 5 - e.g., 0.2 mol] \\
\textit{(Risposta corretta: [Index for 0.02 mol])}

\vspace{0.5em}
\textbf{Question (English Translation):} How many moles correspond to 5 mL (d=1.8 g$\cdot$cm$^{-3}$) of a compound having a molar mass of 450 g$\cdot$mol$^{-1}$? \\
\textbf{Possible answers:} \\
1. [Option 1] \\
2. [Option 2] \\
3. [Option 3] \\
4. [Option 4] \\
5. [Option 5] \\
\textit{(Correct Answer: [Index for 0.02 mol])}
\end{quote}

\subsection{Mathematics Example}
\begin{quote}
\small
\textbf{Domanda:} Dati tre segmenti AA', BB' e CC' tali che: AA' = 2 cm, BB' = 1,5 * AA', CC' = 2,0 * BB'. Quale triangolo è possibile costruire con questi lati? \\
\textbf{Possibili risposte:} \\
1. Non è possibile costruire nessun triangolo \\
2. Un triangolo rettangolo \\
3. Un triangolo ottusangolo \\
4. Un triangolo scaleno \\
5. Un triangolo acutangolo \\
\textit{(Risposta corretta: 1)}

\vspace{0.5em}
\textbf{Question (English Translation):} Given three segments AA', BB', and CC' such that: AA' = 2 cm, BB' = 1.5 * AA', CC' = 2.0 * BB'. Which triangle is possible to construct with these sides? \\
\textbf{Possible answers:} \\
1. It is not possible to construct any triangle \\
2. A right-angled triangle \\
3. An obtuse-angled triangle \\
4. A scalene triangle \\
5. An acute-angled triangle \\
\textit{(Correct Answer: 1)}
\end{quote}

\section{Per-Subject Model Performance}
\label{app:subject_results}

\begin{table}[h]
    \centering
    \caption{Per-subject accuracy (\%) on MedBench-IT for Standard (Std.) and Reasoning (Reas.) prompts. Models sorted as in Table 3.}
    \label{tab:subject_results}
    \resizebox{1.79\columnwidth}{!}{%
    \begin{tabular}{l rr rr rr rr rr rr}

        \toprule
        \multirow{2}{*}{\textbf{Model}} & \multicolumn{2}{c}{\textbf{Biology}} & \multicolumn{2}{c}{\textbf{Chemistry}} & \multicolumn{2}{c}{\textbf{Gen. Culture}} & \multicolumn{2}{c}{\textbf{Physics}} & \multicolumn{2}{c}{\textbf{Logic}} & \multicolumn{2}{c}{\textbf{Math}} \\
        \cmidrule(lr){2-3} \cmidrule(lr){4-5} \cmidrule(lr){6-7} \cmidrule(lr){8-9} \cmidrule(lr){10-11} \cmidrule(lr){12-13}
        & \textbf{Std.} & \textbf{Reas.} & \textbf{Std.} & \textbf{Reas.} & \textbf{Std.} & \textbf{Reas.} & \textbf{Std.} & \textbf{Reas.} & \textbf{Std.} & \textbf{Reas.} & \textbf{Std.} & \textbf{Reas.} \\
        \midrule
        \multicolumn{13}{l}{\textit{API-based Models}} \\
        DeepSeek-R1 & 93.8 & 93.8 & 94.7 & 94.6 & 91.1 & 91.4 & 94.3 & 94.3 & 85.0 & 84.8 & 90.8 & 90.5 \\
        o1-preview & 93.7 & 93.7 & 92.8 & 93.3 & 90.7 & 91.4 & 89.2 & 90.1 & 78.5 & 80.8 & 84.1 & 87.2 \\
        Claude 3.5 Son. & 92.2 & 92.6 & 91.5 & 92.0 & 89.9 & 90.0 & 90.4 & 89.2 & 75.1 & 76.8 & 83.3 & 84.8 \\
        DeepSeek Chat & 91.8 & 91.8 & 89.8 & 89.6 & 87.4 & 88.0 & 89.8 & 89.6 & 70.9 & 74.2 & 83.9 & 85.8 \\
        GPT-4o & 92.4 & 92.4 & 88.0 & 88.3 & 87.3 & 88.5 & 86.4 & 88.9 & 64.9 & 74.0 & 76.2 & 82.2 \\
        GPT-4 Turbo & 90.5 & 87.3 & 86.9 & 82.8 & 85.6 & 83.8 & 86.7 & 80.5 & 65.3 & 58.7 & 79.0 & 72.3 \\
        Claude 3.5 Haiku & 86.7 & 86.3 & 84.5 & 83.8 & 85.1 & 84.2 & 83.3 & 82.5 & 62.7 & 60.1 & 74.9 & 73.0 \\
        GPT-4o mini & 87.4 & 88.0 & 81.4 & 82.3 & 81.5 & 82.9 & 81.4 & 83.4 & 58.7 & 64.1 & 76.2 & 78.4 \\
        GPT-3.5 Turbo & 71.3 & 71.9 & 44.8 & 46.1 & 68.2 & 68.3 & 32.4 & 34.0 & 29.8 & 33.1 & 21.1 & 25.1 \\
        \multicolumn{13}{l}{\textit{Local Models (<30B)}} \\
        Phi-4 & 87.6 & 81.0 & 80.3 & 70.0 & 78.8 & 72.8 & 76.8 & 60.0 & 57.8 & 46.7 & 68.2 & 54.0 \\
        Qwen 2.5 14B & 81.2 & 83.8 & 73.5 & 80.3 & 76.6 & 81.4 & 72.8 & 76.7 & 56.3 & 62.1 & 69.5 & 71.7 \\
        Lexora Med. 7B & 76.3 & 78.9 & 62.4 & 69.1 & 67.1 & 70.4 & 56.8 & 61.1 & 42.0 & 44.5 & 54.0 & 57.0 \\
        Gemma 2 9B & 79.2 & 81.4 & 61.6 & 69.3 & 70.8 & 73.1 & 52.5 & 60.4 & 41.0 & 46.9 & 44.3 & 54.3 \\
        Qwen 2.5 7B & 74.2 & 78.9 & 59.2 & 67.9 & 67.9 & 71.4 & 57.9 & 63.8 & 42.3 & 45.9 & 54.3 & 59.9 \\
        Llama 3.1 8B & 63.4 & 70.9 & 47.1 & 55.4 & 66.4 & 69.0 & 41.8 & 48.3 & 34.4 & 39.0 & 34.2 & 41.9 \\
        Maestrale v0.4 & 68.1 & 68.3 & 48.3 & 50.1 & 67.8 & 67.7 & 39.5 & 41.4 & 31.8 & 34.0 & 27.8 & 32.1 \\
        Aya Expanse 8B & 61.8 & 0.4 & 43.3 & 0.2 & 64.4 & 0.1 & 37.4 & 0.1 & 29.5 & 0.1 & 26.4 & 0.3 \\
        Gemma 2 2B & 52.8 & 43.6 & 38.8 & 31.5 & 49.6 & 40.5 & 35.9 & 28.1 & 28.5 & 23.4 & 29.1 & 23.9 \\
        Qwen 2.5 0.5B & 29.2 & 23.3 & 21.7 & 18.2 & 27.5 & 23.2 & 17.3 & 14.8 & 17.1 & 14.2 & 19.4 & 16.2 \\
        \bottomrule
    \end{tabular}
    }
\end{table}

\end{document}